\documentclass[twoside]{article}
\usepackage[a4paper]{geometry}
\usepackage[latin1]{inputenc} 
\usepackage[T1]{fontenc} 
\usepackage{hyperref}
\usepackage{graphicx, amsmath, theorem, amssymb, algorithmic}

\begin{document}

\newcommand{\proof}{{\bf Proof:~}}
\newcommand{\uio}{$\Sigma^{(0)}$}
\newcommand{\uioo}{$\Sigma^{(1)}$}
\newcommand{\uiok}{$\Sigma^{(k)}$}
\newcommand{\eorc}{$EORC$}
\newcommand{\uiokm}{$\Sigma^{(\overline{k})}$}
\newcommand{\uioK}{$\Sigma^{(k+1)}$}

\title{Closed-form solution to cooperative visual-inertial structure from motion}

\author{Agostino Martinelli\\INRIA, Grenoble, France\\agostino.martinelli@inria.fr}


\maketitle

\begin{abstract}
This paper considers the problem of visual-inertial sensor fusion in the cooperative case and it provides new theoretical contributions, which regard its observability and its resolvability in closed form. The case of two agents is investigated. Each agent is equipped with inertial sensors (accelerometer and gyroscope) and with a monocular camera. By using the monocular camera, each agent can observe the other agent. No additional camera observations (e.g., of external point features in the environment) are considered. All the inertial sensors are assumed to be affected by a bias.
First, the entire observable state is analytically derived. This state includes the absolute scale, the relative velocity between the two agents, the three Euler angles that express the rotation between the two agent frames and all the accelerometer and gyroscope biases.
Second, the paper provides the extension of the closed-form solution given in \cite{IJCV14} (which holds for a single agent)  to the aforementioned cooperative case. The impact of the presence of the bias on the performance of this closed-form solution is investigated. As in the case of a single agent, this performance is significantly  sensitive to the presence of a bias on the gyroscope, while, the presence of a bias on the accelerometer is negligible.
Finally, a simple and effective method to obtain the gyroscope bias is proposed.
Extensive simulations clearly show that the proposed method is successful. It is amazing that, it is possible to automatically retrieve the absolute scale and simultaneously calibrate the gyroscopes not only without any prior knowledge (as in \cite{Kai16}), but also without external point features in the environment.
\end{abstract}

\newpage

\tableofcontents

\newpage

\section{Introduction}\label{SectionIntro}

The problem of fusing visual and inertial data has been extensively investigated in the past (e.g., \cite{Arm07,Fors14,Hua09,Jon11,Li13}). 
Recently, this sensor fusion problem has been successfully
addressed by enforcing observability constraints \cite{Hesc14,Hua15},
and by using optimization-based approaches \cite{Fors15,Hua11,Inde13,Leute14,Lup12,Mou07,Mou08}. 
These optimization methods outperform filter-based algorithms in
terms of accuracy due to their capability of relinearizing past
states. On the other hand, the optimization process can be
affected by the presence of local minima. For this reason, a deterministic solution able to automatically determine the state without initialization has been introduced \cite{Kai16,TRO12,IJCV14}.

Visual and inertial sensors have also been used in a cooperative scenario (e.g., for cooperative mapping in \cite{Guo16}). However, in the cooperative case, a deterministic solution able to automatically determine the state without initialization (as in \cite{Kai16,TRO12,IJCV14}), still misses.
The goal of this paper is precisely to address this lack. We relax the assumption that one or more features are available from the environment: we investigate the extreme case where no point features are available. Additionally, we consider the critical case of only two agents. In other words, we are interested in investigating the minimal case. If we prove that the absolute scale is observable, we can conclude that it is observable in all the other cases. Each  agent is equipped with an Inertial Measurement Unit (IMU) and a monocular camera. By using the monocular camera, each  agent can observe the other one. Note that, we do not assume that these camera observations contain metric information (due for instance to the known size of the observed  agent). The two agents can operate far from each other and a single camera observation only consists of the bearing of the observed  agent in the frame of the observer. In other words, each  agent acts as a moving point feature with respect to the other  agent. The system is defined in section \ref{SectionSystem}.

%

The first questions we wish to answer are: {\it Is it possible to retrieve the absolute scale in these conditions? And the absolute roll and pitch angles?} More in general, we want to determine the entire observable state, i.e., all the physical quantities that it is possible to determine by only using the information contained in the sensor data (from the two cameras and the two IMUs) during a short time interval. In \cite{MRS17} we provided the answers to these questions in the case when the inertial measurements are unbiased. These results are summarized in the first part of section \ref{SectionObservableState}. Then, in the second part of this section, we provide a full answer even in presence of biased measurements (both the ones from the accelerometers and the ones from the gyroscopes).

Section \ref{SectionCFS} provides a closed-form solution, able to obtain the observable state by only using visual and inertial measurements from the two agents delivered during a short time interval. This is precisely the extension of the closed form solution in \cite{TRO12,IJCV14} to the cooperative case.
Then, the paper demonstrates the efficiency of this solution.
By nature, a closed-form solution is deterministic and,
thus, does not require any initialization.
We perform
simulations with plausible MAV motions and synthetic noisy
sensor data (section \ref{SectionLimitation}). This allows us to identify limitations
of the solution and bring modifications to overcome them. 
In practice, we perform exactly the same investigation done in \cite{Kai16} for the case of a single agent.
Specifically, we investigate the impact of biased inertial measurements.
We show that a large bias on the
accelerometer does not significantly worsen the performance (section \ref{SubSectionPerformanceBIASAcc}).
One major limitation is the impact of biased gyroscope
measurements (section \ref{SubSectionPerformanceBIASGyro}). In other words, the performance becomes very
poor in presence of a bias on the gyroscopes of the two agents and, in practice,
the overall method can only be successfully used with
very precise - and expensive - gyroscopes. In section \ref{SectionBiasCalibration}, we introduce
a simple method that automatically estimates both these biases. By
adding this new method for the bias estimation to the solution presented in section \ref{SectionCFS}, we obtain results that are equivalent to the ones
in absence of bias (section \ref{SectionExperiments}). 

This method is suitable for MAVs that operate in absence of point features and need to recover the absolute scale in few seconds. The solution does not need initialization. Additionally,  is robust to the bias and
automatically calibrates the gyroscopes.

\section{The system}\label{SectionSystem}

We consider two agents that move in a $3D-$environment. Each agent is equipped with an Inertial Measurement Unit (IMU), which consists of three orthogonal accelerometers and three orthogonal  gyroscopes. Additionally, each agent is equipped with a monocular camera. We assume that, for each agent, all the sensors share the same frame. Without loss of generality, we define the agent local frame as this common frame. The accelerometer sensors perceive both the gravity and the inertial acceleration in the local frame. The gyroscopes provide the angular speed in the local frame. Finally, the monocular camera mounted on the first agent (or the second agent) provides the bearing of the second agent (or the first agent) in its local frame.


\begin{figure}[htbp]
\begin{center}
\includegraphics[width=.9\columnwidth]{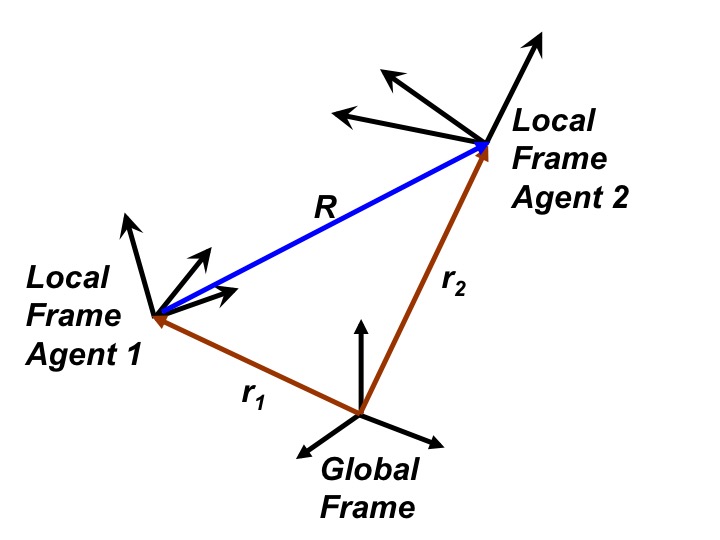}
\caption{The global frame and the two local frames (attached to the first and to the second agent, respectively). $r_1$ and $r_2$ are their position, expressed in the global frame. $R$ is the relative position of the second agent in the local frame of the first agent.} \label{Fig2Vehicles}
\end{center}
\end{figure}

\noindent We adopt the following notations:

\begin{itemize}

\item $R$ is the position of the second agent in the local frame of the first agent;

\item $V$ is the relative velocity of the second agent with respect to the first agent, expressed in the frame of the first agent (note that this velocity is not simply the time derivative of $R$ because of the rotations accomplished by the first agent);

\item $O$ is the orthonormal matrix that characterizes the rotation between the two local frames; specifically, 
for a vector with given coordinates in the local frame of the second agent, we obtain its coordinates in the local frame of the first agent by pre multiplying by $O$;

\end{itemize}

\noindent Additionally, we denote by $A^1,~\Omega^1$ the acceleration and the angular speed of the first agent expressed in  the first agent local frame. Similarly, we denote by $A^2,~\Omega^2$ the acceleration and the angular speed of the second agent expressed in the second agent local frame. Regarding the accelerations, they include both the inertial acceleration and the gravity.
We denote by $B_A^1,~B_A^2, ~B_\Omega^1$ and $B_\Omega^2$ the biases that affect the measurements from the two accelerometers and the two gyroscopes.

\noindent We conclude this section by providing the analytic expressions of the two camera observations. The first camera provides the vector $R$, up to a scale. The scale is precisely the distance  between the first agent and the second agent at the time of the camera measurement. The second camera provides the vector $-O^TR$, up to a scale. Note that, in the special case when the two cameras are synchronized, the scale coincides.

\section{Observable state and its dynamics}\label{SectionObservableState}

\subsection{The unbiased case}

In appendix \ref{AppendixOASystem} we analyzed the observability properties of the system defined in section \ref{SectionSystem}, in the case when the inertial measurements are unbiased. We showed that all the observable states are:

\begin{itemize}

\item the three components of $R$;

\item the three components of $V$;

\item the three Euler angles that uniquely define the matrix $O$.

\end{itemize}

\noindent Basically, this result tells us that, starting from the measurements delivered by the two IMUs and the two cameras during a given time interval, we can reconstruct the previous three quantities. Estimating other physical quantities (e.g., the absolute roll and pitch angles of the first agent or the second agent) is not possible. Note that we can retrieve the absolute scale, even when no feature is available in the environment. In appendix \ref{AppendixOASystem} we also proven that, the same observable state, characterizes the case when only one of the agents is equipped with a camera. Namely,
the presence of two cameras does not improve the observability properties with respect to the case of a single camera mounted on one of the two agents.
In appendix \ref{AppendixBasicEquations} we derived the analytic expression of the dynamics of the observable states. They are:

\begin{equation}\label{EquationObsDynamicsNoq}
\left[\begin{aligned}
\dot{R} &= \left[ \Omega^1\right]_{\times} R + V \\
\dot{V} &= \left[ \Omega^1\right]_{\times} V + O A^2 - A^1\\
\dot{O} &= \left[ \Omega^1\right]_{\times}^T O + O \left[ \Omega^2\right]_{\times}\\
\end{aligned}
\right.
\end{equation}

\noindent where $\left[ \Omega^{1/2}\right]_{\times}$ are the skew-symmetric matrices associated to the vectors $\Omega^{1/2}$:

\begin{equation}\label{EquationOmegaMatrix}
\left[ \Omega^i\right]_{\times}=
\left[
\begin{array}{ccc}
0 & \Omega_z^i & -\Omega_y^i\\
-\Omega_z^i & 0 &  \Omega_x^i\\
 \Omega_y^i & -\Omega_x^i & 0\\
\end{array}
\right],
~~~
i=1,2
\end{equation}

\noindent Finally, the two cameras provide the two vectors, $R$ and $-O^TR$, up to a scale.

\subsection{The biased case}

The goal of this subsection is to obtain the observable state when the inertial measurements are corrupted by the biases $B_A^1,~B_A^2, ~B_\Omega^1$ and $B_\Omega^2$. Since the presence of the bias cannot improve the observability properties, we characterize our system by including in the observable state that holds in absence of bias (i.e., the state mentioned in the previous subsection), all the $12$ components of the $4$ bias vectors. If we prove that this state is observable, we can conclude that it is the entire observable state, i.e., any other physical quantity independent from its components is unobservable. Additionally, we will consider the case when only the first agent is equipped with a camera. Again, by proving that in these conditions the previous state is observable, we can conclude that the same observable state characterizes the case of two cameras.

The bias is time dependent. However, it changes very slowly with time. In particular, in the interval of few seconds, it can be considered constant. Since we will consider time intervals no longer than $4$ seconds (and, as it will be shown, this will allow us to auto calibrate the inertial sensors with very high accuracy), we can assume that the bias is constant during the considered time interval (the same assumption is made in \cite{Kai16}).

We characterize our system by the following state:

\begin{equation}\label{EquationState}
X=[R~V~q~B_\Omega^1~B_A^1~B_\Omega^2~B_A^2]^T
\end{equation}

\noindent where $q$ is the unit quaternion associated to the rotation matrix $O$.

The dimension of this state is equal to $22$. Actually, the components of this state are not independent, since $q$ is a unit quaternion. In other words, we have: $(q_t)^2+(q_x)^2+(q_y)^2+(q_z)^2=1$.

The dynamics of the state defined in (\ref{EquationState}) are given by the following equations:

\begin{equation}\label{EquationDynamicsBIAS}
\left[\begin{aligned}
\dot{R}~~ &= \left[ \Omega'^1\right]_{\times} R + V \\
\dot{V}~~&= \left[ \Omega'^1\right]_{\times} V + O A'^2 - A'^1\\
\dot{q}~~&= -\frac{1}{2}\Omega'^1_q q + \frac{1}{2} q\Omega'^2_q\\
\dot{B}_\Omega^1 &= \dot{B}_A^1= \dot{B}_\Omega^2= \dot{B}_A^2=0\\
\end{aligned}
\right.
\end{equation}

\noindent where:
\begin{itemize}

\item $\Omega'^1=\Omega^1+B_\Omega^1$, $A'^1=A^1+B_A^1$, $\Omega'^2=\Omega^2+B_\Omega^2$, $A'^2=A^2+B_A^2$.

\item The matrix $O$, can be uniquely expressed in terms of the components of the quaternion $q$.

\item $\Omega'^1_q$ is the imaginary quaternion associated with $\Omega'^1$, i.e.,: $\Omega'^1_q=0+\Omega'^1_x i+\Omega'^1_y j +\Omega'^1_z k$. The same holds for $\Omega'^2_q$

\end{itemize}

\noindent The observation functions are the two scalar functions $h_u ~h_v$:

\begin{equation}\label{EquationFirstCamera}
h\triangleq  [h_u,~h_v]^T=\left[
\frac{R_x}{R_z},
~\frac{R_y}{R_z} \right]^T
\end{equation}

\noindent Additionally, we need to add the observation function that expresses the constraint that  $q$ is a unit quaternion. The additional observation is:

\begin{equation}\label{EquationQUnit}
h_{const}(X)\triangleq  (q_t)^2+(q_x)^2+(q_y)^2+(q_z)^2
\end{equation}

\noindent The analytic derivation of this system observability is provided in appendix \ref{AppendixObservabilityBIAS}.
We summarize its result:

{\bf For the system defined in section \ref{SectionSystem} the observable state is the one given in (\ref{EquationState}). This holds both in the case when both the agents are equipped with a camera and in the case when only one agent is equipped with a camera. Additionally, the observable state remains the same even in the case when the camera is a linear camera, i.e., it only provides the azimuth of the other agent in its local frame.}

\section{Closed-form solution}\label{SectionCFS}

In this section we will refer to the two agents as to two rigid bodies and we denote them by $\mathcal{B}_1$ and $\mathcal{B}_2$. This is to emphasize the generality of the closed form solution here introduced, which is not limited to the framework of aerial navigation. 

\noindent Let us consider a given time interval $(t_A, ~t_B)$. 
Let us denote by $R_A$, $V_A$ and $O_A$, the values of $R$, $V$ and $O$ at time $t_A$. 
The goal of this section is to obtain these values in closed form, only in terms of the measurements provided during the considered time interval (both from the cameras and the two IMUs).
Note that, the length of the considered time interval (i.e., $t_B-t_A$) is very small (about $2$, $3$ seconds).

In section \ref{SectionSolution} we directly provide the solution. All the analytic derivations are given later, in section \ref{SubSectionDerivation}.

\subsection{The solution}\label{SectionSolution}

We distinguish the case when only $\mathcal{B}_1$ is equipped with a camera, from the case when both $\mathcal{B}_1$ and $\mathcal{B}_2$ are equipped with a camera and the observations are synchronized. In particular, we assume that, during our time interval, the camera (or the two cameras) performs $n$ observations at the times $t_j$, ($j=1,\cdots,n$), with $t_1=t_A$ and $t_n=t_B$.
In both cases (single camera and two cameras synchronized), we will obtain the components of $R_A$, $V_A$ and $O_A$ by simply solving the linear system:

\begin{equation}\label{EquationLinearSystem}
\Xi x=b
\end{equation}

\noindent where:

\begin{itemize}

\item $\Xi$ is a matrix with dimension $3n\times (15+n)$ in the case of a single camera and $6n\times (21+n)$ in the case of two cameras synchronized.

\item $x$ is the vector that contains all the unknowns; in the case of a single camera it contains the components of $R_A$, $V_A$ and $O_A$ and the $n$ distances between $\mathcal{B}_1$ and $\mathcal{B}_2$ at the times $t_1,t_2,\cdots, t_n$; in the case of two cameras synchronized, it also contains the components of the two vectors $O_AR_A$ and $O_AV_A$.

\item $b$ is a vector with dimension $3n$ in the case of a single camera and $6n$ in the case of two cameras synchronized.

\end{itemize}

\noindent In the next two subsections we provide the expressions of the vector $x$, the matrix $\Xi$ and the vector $b$, in the case of a single camera and two synchronized cameras, respectively. Then, in section \ref{SubSectionDerivation}, we provide all the analytic steps to obtain these expressions and the linear system in (\ref{EquationLinearSystem}).

\subsubsection{Single camera}\label{SubSubSectionSingleCameraSolution}

The vector $x$, i.e., the vector that includes all the unknowns is:

\begin{equation}\label{Equationx}
x \equiv 
[R_A^T, V_A^T, O_{A_{11}}, O_{A_{21}}, O_{A_{31}}, O_{A_{12}}, O_{A_{22}}, O_{A_{32}}, O_{A_{13}}, O_{A_{23}}, O_{A_{33}},\lambda_1,\cdots,\lambda_n]^T
\end{equation}

\noindent where $\lambda_1,\cdots,\lambda_n$, are the distances between $\mathcal{B}_1$ and $\mathcal{B}_2$ at the times $t_1,\cdots,t_n$. Hence, $x$ contains the components of $R_A$, the components of $V_A$, all the entries of the matrix $O_A$ and all the aforementioned distances. Since $O_A$ is orthonormal, by including all its entries in $x$ we are ignoring six quadratic constraints. 
The matrix $\Xi$ is:

\begin{equation}\label{EquationA}
\Xi \equiv
\end{equation}
\[
\left[
\begin{array}{c|c|c|c|c|c|c|c|c|c|c|c}
I_3&0_{3 3} &0_{3 3}&0_{3 3}&0_{3 3}&-\mu_1&0_3&\cdots&\cdots&\cdots&\cdots&0_3\\
I_3&\Delta_2I_3&\beta^2_{x2}I_3&\beta^2_{y2}I_3&\beta^2_{z2}I_3&0_3&-\mu_2&0_3&\cdots&\cdots&\cdots&0_3\\
I_3&\Delta_3I_3&\beta^2_{x3}I_3&\beta^2_{y3}I_3&\beta^2_{z3}I_3&0_3&0_3&-\mu_3&0_3&\cdots&\cdots&0_3\\
\cdots&\cdots&\cdots&\cdots&\cdots&\cdots&\cdots&\cdots&\cdots&\cdots&\cdots&\cdots\\
I_3&\Delta_jI_3&\beta^2_{xj}I_3&\beta^2_{yj}I_3&\beta^2_{zj}I_3&0_3&\cdots&0_3&-\mu_j&0_3&\cdots&0_3\\
\cdots&\cdots&\cdots&\cdots&\cdots&\cdots&\cdots&\cdots&\cdots&\cdots&\cdots&\cdots\\
I_3&\Delta_{n-1}I_3&\beta^2_{x~n-1}I_3&\beta^2_{y~n-1}I_3&\beta^2_{z~n-1}I_3&0_3&\cdots&\cdots&\cdots&0_3&-\mu_{n-1}&0_3\\
I_3&\Delta_nI_3&\beta^2_{xn}I_3&\beta^2_{yn}I_3&\beta^2_{zn}I_3&0_3&\cdots&\cdots&\cdots&\cdots&0_3&-\mu_n\\
\end{array}
\right]
\]

\noindent where:

\begin{itemize}

\item $I_3$ is the identity $3\times 3$ matrix, $0_{3 3}$ is the $3\times 3$ zero matrix, $0_3$ the zero $3\times 1$ vector;

\item $\mu_1,\cdots,\mu_n$ are the unit vectors provided by the camera (i.e., the directions of $\mathcal{B}_2$ in the frame of $\mathcal{B}_1$ at times $t_1,\cdots,t_n$) rotated by pre-multiplying them by the matrix $M^1(t_j)^T$ (this matrix is defined in section \ref{SubSectionDerivation} by equation (\ref{EquationM1M2}) and can be obtained by only using the gyroscope measurements from the IMU of $\mathcal{B}_1$ in the considered time interval);

\item $\Delta_j\equiv t_j-t_1=t_j-t_A$ ($j=2,\cdots,n$);

\item $\beta^2_{xj}$, $\beta^2_{yj}$ and $\beta^2_{zj}$ ($j=2,\cdots,n$) are defined in section \ref{SubSectionDerivation} and can be obtained by only using the gyroscope and accelerometer measurements from the IMU of $\mathcal{B}_2$ in the time interval $(t_A,~t_j)$.

\end{itemize}

\noindent The vector $b$ is:

\begin{equation}\label{Equationb}
b \equiv 
[\beta^{1~T}_1,\beta^{1~T}_2,\cdots,\beta^{1~T}_j,\cdots,\beta^{1~T}_n]^T
\end{equation}

\noindent where $\beta^1_j$ ($j=2,\cdots,n$) are defined in section \ref{SubSectionDerivation} and can be obtained by only using the gyroscope and accelerometer measurements from the IMU of $\mathcal{B}_1$ in the time interval $(t_A,~t_j)$.

\subsubsection{Two cameras synchronized}\label{SubSubSectionTwoCamerasSolution}

The vector $x$ also includes the components of the two vectors $O_AR_A$ and $O_AV_A$. These vectors are actually dependent on the remaining components of $x$. Since we obtain the solution by inverting the linear system in (\ref{EquationLinearSystem}), we are ignoring six further quadratic equations. The vector $x$ is:

\begin{equation}\label{Equationx2}
x \equiv 
[R_A^T, V_A^T, (O_AR_A)^T, (O_AV_A)^T, O_{A_{11}}, O_{A_{21}}, O_{A_{31}}, O_{A_{12}}, O_{A_{22}}, O_{A_{32}}, O_{A_{13}}, O_{A_{23}}, O_{A_{33}},\lambda_1,\cdots,\lambda_n]^T
\end{equation}

\noindent The matrix $\Xi$ is:

\begin{equation}\label{EquationA2}
\Xi \equiv
\end{equation}
{\tiny
\[
\left[
\begin{array}{c|c|c|c|c|c|c|c|c|c|c|c|c|c}
I_3&0_{3 3} &0_{3 3} &0_{3 3} &0_{3 3}&0_{3 3}&0_{3 3}&-\mu_1&0_3&\cdots&\cdots&\cdots&\cdots&0_3\\
0_{3 3} &0_{3 3} &I_3&0_{3 3} &0_{3 3}&0_{3 3}&0_{3 3}&-\nu_1&0_3&\cdots&\cdots&\cdots&\cdots&0_3\\
I_3&\Delta_2I_3&0_{3 3} &0_{3 3} &\beta^2_{x2}I_3&\beta^2_{y2}I_3&\beta^2_{z2}I_3&0_3&-\mu_2&0_3&\cdots&\cdots&\cdots&0_3\\
0_{3 3} &0_{3 3} &I_3&\Delta_2I_3&[\beta^1_2]^{up}&[\beta^1_2]^{center}&[\beta^1_2]^{down}&0_3&-\nu_2&0_3&\cdots&\cdots&\cdots&0_3\\
I_3&\Delta_3I_3&0_{3 3} &0_{3 3} &\beta^2_{x3}I_3&\beta^2_{y3}I_3&\beta^2_{z3}I_3&0_3&0_3&-\mu_3&0_3&\cdots&\cdots&0_3\\
0_{3 3} &0_{3 3} &I_3&\Delta_3I_3&[\beta^1_3]^{up}&[\beta^1_3]^{center}&[\beta^1_3]^{down}&0_3&0_3&-\nu_3&0_3&\cdots&\cdots&0_3\\
\cdots&\cdots&\cdots&\cdots&\cdots&\cdots&\cdots&\cdots&\cdots&\cdots&\cdots&\cdots&\cdots&\cdots\\
I_3&\Delta_jI_3&0_{3 3} &0_{3 3}&\beta^2_{xj}I_3&\beta^2_{yj}I_3&\beta^2_{zj}I_3&0_3&\cdots&0_3&-\mu_j&0_3&\cdots&0_3\\
0_{3 3} &0_{3 3} &I_3&\Delta_jI_3&[\beta^1_j]^{up}&[\beta^1_j]^{center}&[\beta^1_j]^{down}&0_3&\cdots&0_3&-\nu_j&0_3&\cdots&0_3\\
\cdots&\cdots&\cdots&\cdots&\cdots&\cdots&\cdots&\cdots&\cdots&\cdots&\cdots&\cdots&\cdots&\cdots\\
I_3&\Delta_{n-1}I_3&0_{3 3} &0_{3 3}&\beta^2_{x~n-1}I_3&\beta^2_{y~n-1}I_3&\beta^2_{z~n-1}I_3&0_3&\cdots&\cdots&\cdots&0_3&-\mu_{n-1}&0_3\\
0_{3 3} &0_{3 3} &I_3&\Delta_{n-1}I_3&[\beta^1_{n-1}]^{up}&[\beta^1_{n-1}]^{center}&[\beta^1_{n-1}]^{down}&0_3&\cdots&\cdots&\cdots&0_3&-\nu_{n-1}&0_3\\
I_3&\Delta_nI_3&0_{3 3} &0_{3 3} &\beta^2_{xn}I_3&\beta^2_{yn}I_3&\beta^2_{zn}I_3&0_3&\cdots&\cdots&\cdots&\cdots&0_3&-\mu_n\\
0_{3 3} &0_{3 3} &I_3&\Delta_nI_3&[\beta^1_n]^{up}&[\beta^1_n]^{center}&[\beta^1_n]^{down}&0_3&\cdots&\cdots&\cdots&\cdots&0_3&-\nu_n\\
\end{array}
\right]
\]
}

\noindent where:

\begin{itemize}

\item $\nu_1,\cdots,\nu_n$ are the unit vectors provided by the second camera (i.e., the directions of $\mathcal{B}_1$ in the frame of $\mathcal{B}_2$ at times $t_1,\cdots,t_n$) rotated by pre-multiplying them by the matrix $M^2(t_j)^T$  (this matrix is defined in section \ref{SubSectionDerivation} by equation (\ref{EquationM1M2}) and can be obtained by only using the gyroscope measurements from the IMU of $\mathcal{B}_2$ in the time interval $(t_A,~t_j)$);

\item $[\beta^1_j]^{up} \equiv\left[
\begin{array}{ccc}
\beta^1_{xj}&\beta^1_{yj}&\beta^1_{zj}\\
0&0&0\\
0&0&0\\
\end{array}
\right]
$, 
$[\beta^1_j]^{center} \equiv\left[
\begin{array}{ccc}
0&0&0\\
\beta^1_{xj}&\beta^1_{yj}&\beta^1_{zj}\\
0&0&0\\
\end{array}
\right]
$
and
$[\beta^1_j]^{down} \equiv\left[
\begin{array}{ccc}
0&0&0\\
0&0&0\\
\beta^1_{xj}&\beta^1_{yj}&\beta^1_{zj}\\
\end{array}
\right]
$, 
$j=2,\cdots,n$

\end{itemize}

\noindent The vector $b$ is:

\begin{equation}\label{Equationb2}
b \equiv 
[\beta^{1~T}_1,\beta^{2~T}_1,\beta^{1~T}_2,\beta^{2~T}_2,\cdots,\beta^{1~T}_j,\beta^{2~T}_j,\cdots,\beta^{1~T}_n,\beta^{2~T}_n]^T
\end{equation}


We conclude this section by remarking that all the components of the matrix $\Xi$ and the vector $b$ depend only on the measurements from the IMUs and the camera (or the two cameras) delivered during the time interval $(t_A, ~t_B)$. 
As a result, the solution is able to obtain the entire observable state without any prior knowledge (e.g., initialization). In particular, it provides the state as a simple expression of the measurements delivered during the time interval $(t_A, ~t_B)$. In addition, since the time interval is very short ($2,3$ seconds), the solution is also drift-free.

Note that we include in $x$ all the entries of the matrix $O_A$ and, in the case of two synchronized cameras, we also include the components of the two vectors $O_AR_A$ and $O_AV_A$.
This means that by obtaining the observable state through the inversion of the linear system in (\ref{EquationLinearSystem}), we are considering independent the entries of $O_A$ and, in the case of two cameras synchronized, the entries of $O_AR_A$ and $O_AV_A$ independent from the remaining components of $x$. The fact that the matrix $O_A$ is orthonormal, means that we are ignoring $6$ quadratic equations, i.e., the equations that express the fact that each column of the matrix is a unit vector (3 equations) and that the three columns are orthogonal one each other (3 equations). In the case of two cameras synchronized, including the components of the two vectors $O_AR_A$ and $O_AV_A$ in $x$, means that we are ignoring six further quadratic equations. All these quadratic equations could be exploited in a second step to improve the precision (e.g., by minimizing the cost function defined as the norm of the residual of both the linear system in (\ref{EquationLinearSystem}) and the system of the aforementioned quadratic equations).

\subsection{Analytic Derivation}\label{SubSectionDerivation}

\begin{figure}[htbp]
\begin{center}
\includegraphics[width=1\columnwidth]{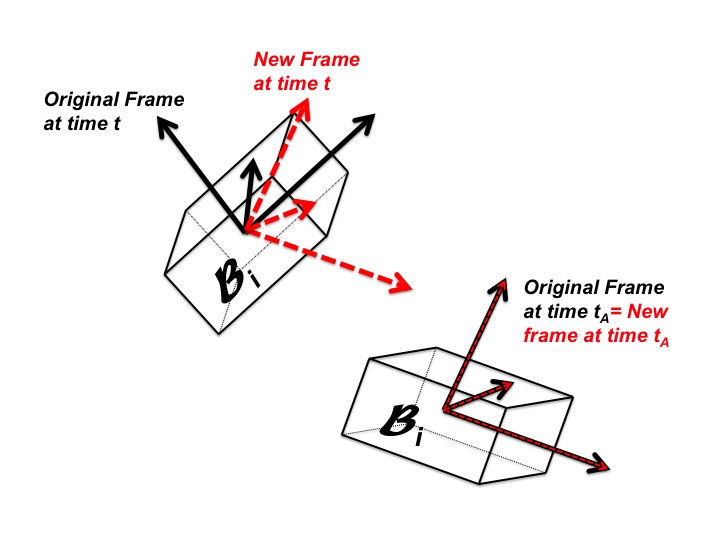}
\caption{Original and new local frame of $\mathcal{B}_i$ ($i=1,2$). The original frame is attached to the rigid body and rotates with it. At time $t_A$ the two frames coincide. The new frame does not rotate and its origin coincides with the origin of the original frame at any time.} \label{Fig2Frames}
\end{center}
\end{figure}

We start our derivation by introducing a new local frame for each rigid body (i.e., one new frame for $\mathcal{B}_1$ and one new frame for $\mathcal{B}_2$). Each new frame is defined as follows. It shares the same origin with the original local frame. Additionally, it does not rotate and its orientation coincides with the one of the original frame at the time $t_A$. From now on, we will refer to this frame as to the {\it new} frame. Additionally, we will refer to the original local frame, namely the one defined at the beginning of section \ref{SectionSystem}, as to the {\it original} frame. Figure \ref{Fig2Frames} illustrates the original and the new frame of $\mathcal{B}_i$ ($i=1,2$). Specifically, in this figure, the considered rigid body accomplishes a translation and rotation between time $t_A$ and $t$. The original frame is in black. The new frame is in red dashed line. The two frames coincide at time $t_A$.

Let us introduce the following notation:

\begin{itemize}

\item $\xi$ is the position of $\mathcal{B}_2$ in the new local frame of $\mathcal{B}_1$;

\item $\eta$ is the relative velocity of $\mathcal{B}_2$ with respect to $\mathcal{B}_1$, expressed in the new local frame of $\mathcal{B}_1$;

\item $M^1(t)$ is the orthonormal matrix that characterizes the rotation made by $\mathcal{B}_1$ between $t_A$ and $t\in(t_A, ~t_B)$; in other words, it describes the difference in orientation between the original and the new frame of $\mathcal{B}_1$ at a given time $t\in(t_A,~t_B)$;

\item $M^2(t)$ is defined as $M^1(t)$, but for $\mathcal{B}_2$.

\end{itemize}

\noindent By construction we have:

\begin{equation}
\xi_A\equiv \xi(t_A)=R_A~~~~ \eta_A\equiv \eta(t_A)=V_A
\end{equation}

\noindent Additionally, $M^1(t)$ and $M^2(t)$ can be computed by integrating the following first order differential equations:

\begin{equation}\label{EquationM1M2}
\dot{M}_1 = \left[ \Omega^1\right]_{\times}^T M_1~~~\dot{M}_2 = \left[ \Omega^2\right]_{\times}^T M_2
\end{equation}

\noindent with initial conditions: $M_1(t_A)=M_2(t_A)=I_3$, ($I_3$ is the $3\times 3$ identity matrix) and $ \left[ \Omega^1\right]_{\times}$ and $ \left[ \Omega^2\right]_{\times}$ are the matrices defined in (\ref{EquationOmegaMatrix}). Note that, since $t_B-t_A$ does not exceed $3$ seconds, these two matrices can be obtained with very high accuracy by using the measurements from the gyroscopes delivered in the considered time interval. In particular, the drift due to the noise in the gyroscope measurements, is negligible. Regarding the bias, we will show that it can be removed (BIAS).

From (\ref{EquationObsDynamicsNoq}) we have the following dynamics in the new coordinates:

\begin{equation}\label{EquationDynamicsCF}
\left[\begin{aligned}
\dot{\xi} ~&= \eta \\
\dot{\eta} ~&= O_A \mathcal{A}^2 - \mathcal{A}^1\\
\dot{O}_A &=0
\end{aligned}
\right.
\end{equation}

\noindent where:

\begin{itemize}

\item $\mathcal{A}^1$ is the acceleration (gravitational and inertial) of $\mathcal{B}_1$ expressed in the first new local frame (i.e., $\mathcal{A}^1=M^1A^1$);

\item similarly, $\mathcal{A}^2=M^2A^2$ .

\end{itemize}

\noindent Let us introduce the following notation:

\begin{itemize}

\item $w^1$, $w^2$ and $w^3$ are the three columns of the matrix $O_A$, i.e., $O_A=\left[
\begin{array}{ccc}
w^1 & w^2 & w^3\\
\end{array}
\right]
$;

\item $\alpha^1(t)=[\alpha_x^1(t), ~\alpha_y^1(t), ~\alpha_z^1(t)]^T=\int_{t_A}^t \mathcal{A}^1(\tau) d\tau$;

\item $\beta^1(t)=[\beta_x^1(t), ~\beta_y^1(t), ~\beta_z^1(t)]^T=\int_{t_A}^t \alpha^1(\tau) d\tau$;

\item $\alpha^2(t)=[\alpha_x^2(t), ~\alpha_y^2(t), ~\alpha_z^2(t)]^T=\int_{t_A}^t \mathcal{A}^2(\tau) d\tau$;

\item $\beta^2(t)=[\beta_x^2(t), ~\beta_y^2(t), ~\beta_z^2(t)]^T=\int_{t_A}^t \alpha^2(\tau) d\tau$;

\end{itemize}

\noindent Note that all the quantities $\alpha^1(t)$, $\alpha^2(t)$, $\beta^1(t)$ and $\beta^2(t)$ are directly provided by the IMU measurements delivered in the interval $(t_A, ~t)$.

Let us integrate the second equation in (\ref{EquationDynamicsCF}) between $t_A$ and a given $t\in [t_A, t_B]$. We obtain:

\begin{equation}\label{EquationEtaInt}
\eta(t)=\eta_A+w^1 \alpha_x^2(t)+w^2 \alpha_y^2(t)+w^3 \alpha_z^2(t) -\alpha^1(t)
\end{equation}

\noindent and by substituting in the first equation in (\ref{EquationDynamicsCF}) and integrating again, we obtain:

\begin{equation}\label{EquationEtaInt}
\xi(t)=\xi_A + \eta_A (t-t_A)+w^1 \beta_x^2(t)+w^2 \beta_y^2(t)+w^3 \beta_z^2(t) -\beta^1(t)
\end{equation}

\noindent Note that this equation provides $\xi(t)$ as a linear expression of $15$ unknowns, which are the components of the $5$ vectors: $\xi_A$, $\eta_A$, $w^1$, $w^2$ and $w^3$. In the sequel, we build a linear system in these unknowns together with the unknown distances when the cameras perform the measurements. 

\subsubsection{Single camera}\label{SubSubSectionSingleCamera}

The camera on $\mathcal{B}_1$ provides the vector $R(t)=M^1(t)\xi(t)$, up to a scale. We denote by $\lambda(t)$ this scale (this is the distance between $\mathcal{B}_1$ and $\mathcal{B}_2$ at the time $t$). We have $\xi(t)=\lambda(t)\mu(t)$, where $\mu(t)$ is the unit vector with the same direction of $\xi(t)$. Note that our sensors (specifically, the camera together with the gyroscope on $\mathcal{B}_1$) provide precisely the unit vector $\mu(t)$: the camera provides the unit vector along $R(t)$; then, to obtain $\mu(t)$ it suffices to pre multiply this unit vector by $[M^1(t)]^T$.

We remind the reader that the camera performs $n$ observations at the times $t_j$, ($j=1,\cdots,n$), with $t_1=t_A$ and $t_n=t_B$. For notation brevity, for a given time dependent quantity (e.g., $\lambda(t)$), we will denote its value a the time $t_j$ by the pedix $j$ (e.g., $\lambda_j=\lambda(t_j)$).  In this notation, equation (\ref{EquationEtaInt}) becomes:

\begin{equation}\label{EquationCVIsFM1}
\lambda_j\mu_j=\xi_A + \eta_A (t_j-t_A) +w^1 \beta_{xj}^2+w^2 \beta_{yj}^2+w^3 \beta_{zj}^2 -\beta^1_j
\end{equation}

\noindent This is a linear equation in $15+n$ unknowns. The unknowns are:

\begin{itemize}

\item The distances $\lambda_1, \cdots, \lambda_n$.

\item The three components of $\xi_A$.

\item The three components of $\eta_A$.

\item The components of the vectors $w_1,~w_2$ and $w_3$, i.e., the nine entries of the matrix $O_A$.

\end{itemize}

\noindent Note that equation (\ref{EquationCVIsFM1}) is a vector equations, providing $3$ scalar equations. Since this holds for each $j=1,\cdots,n$, we obtain a linear system of $3n$ equations in $15+n$ unknowns. This is precisely the linear system given in (\ref{EquationLinearSystem}) with the vector $x$ given in (\ref{Equationx}), the matrix $A$ given in (\ref{EquationA}) and the vector $b$ given in (\ref{Equationb}).

\subsubsection{Two cameras synchronized}\label{SubSubSectionTwoCameras}

Let us consider now the case when also $\mathcal{B}_2$ is equipped with a camera and, the measurements made by this camera occur at the same times $t_1,\cdots,t_n$ (synchronized cameras). We define $\nu_j$ the unit vector such that 

\[
\lambda_j \nu_j=-O_A^T\xi_j
\]

\noindent Note that $-O_A^T\xi_j$ is the position of $\mathcal{B}_1$ in the {\it new} local frame of $\mathcal{B}_2$. The second camera provides this vector up to a scale and rotated by the rotation defined by the matrix $M^2(t_j)$. Since this matrix is known, we conclude that the camera and the IMU measurements on $\mathcal{B}_2$ provide $\nu_1,\cdots,\nu_n$.

Let us pre multiply both members of (\ref{EquationEtaInt}) by $-O_A^T$. We obtain, for $t=t_j$:

\begin{equation}\label{EquationCVIsFM2}
\lambda_j\nu_j=\xi_A' + \eta_A' (t_j-t_A) -\beta^2_j +v_1^T \beta_{xj}^1+v_2^T \beta_{yj}^1+v_3^T \beta_{zj}^1
\end{equation}

\noindent where:

\begin{itemize}

\item $v_1$, $v_2$ and $v_3$ are the three lines of the matrix $O_A$;

\item $\xi_A'\equiv-O_A^T\xi_A$;

\item $\eta_A'\equiv-O_A^T\eta_A$

\end{itemize}

\noindent Equations (\ref{EquationCVIsFM1}) and (\ref{EquationCVIsFM2}) provide a linear system of $6n$ equations in $21+n$ unknowns. With respect to the case of a single camera, we have $6$ new unknowns (i.e., the components of the two vectors $\xi_A'$ and $\eta_A'$) but we gain $3n$ further equations. The equations in (\ref{EquationCVIsFM1}) for $j=1,\cdots n$ are precisely the linear system in (\ref{EquationLinearSystem}) with the vector $x$ given in (\ref{Equationx2}), the matrix $A$ given in (\ref{EquationA2}) and the vector $b$ given in (\ref{Equationb2}).

\section{Limitations of the Closed-Form Solution}\label{SectionLimitation}

The goal of this section is to find out the limitations of the solution provided in section \ref{SectionCFS} when it is adopted in a real scenario. In particular, special attention will be devoted to the case of an agent equipped with low-cost camera and IMU sensors.
For this reason, this section evaluates the impact of the bias on the performance.

\subsection{Simulation setup}\label{SubSectionSimulationSetup}

We simulate two agents as point particles executing random trajectories.
The trajectories are simulated as follows. Each trial lasts $4~s$. 
The first agent starts at the origin. The second agent starts at the position $(1.0~1.0~1.0)m$.
The initial velocities  are set equal to $(0.1~ -0.1 ~0)ms^{-1}$ and $(0.2~ 0.8 ~0.1)ms^{-1}$, for the first and the second agent, respectively. Finally, the initial orientations are characterized by the following values of the roll, pitch and yaw angles, for the two agents:  $(0.2\pi~-0.3\pi~0.8\pi)rad$ and $(0.2\pi~0.3\pi~-0.8\pi)rad$.
For each trial the agents move randomly. The angular speeds, i.e. 
$\Omega^1$ and $\Omega^2$, are Gaussian. Specifically, their
values at each step of $0.1s$ follow a zero-mean Gaussian distribution
with covariance matrix equal to $(30 deg)^2I_3$, where $I_3$ is the
identity $3\times 3$ matrix. At each time step, the two agent
inertal accelerations are generated as random vectors with
zero-mean Gaussian distribution. In particular, the covariance
matrix of this distribution is set equal to $(1ms^{-2})^2 I_3$.

The agents are equipped with inertial sensors able to
measure at the frequency of $0.5kHz$ the acceleration (the sum of the
gravity and the inertial acceleration) and the angular speed.
These measurements are affected by errors. Specifically, each
measurement is generated by
adding to the true value a random error that follows a Gaussian
distribution. The mean value of this error is zero and
the standard deviation is $0.03ms^{-2}$ for the accelerometer
and $0.1degs^{-1}$ for the gyroscope.
Regarding the camera measurements, they are generated
at a lower frequency. Specifically, the measurements are
generated at $5Hz$. Also these measurements are affected
by errors. Specifically, each measurement is generated by
adding to the true value a random error that follows a zero mean
Gaussian distribution, with variance $1 deg^2$.
All the inertial measurements are corrupted by a bias (see sections \ref{SubSectionPerformanceBIASAcc} and \ref{SubSectionPerformanceBIASGyro}).

We measure our error on the
absolute scale by computing the mean error over all estimated
distances between the agents at times $t_1,\cdots,t_n$. In our case, $n=\frac{4s}{0.2s}=20$. We define the relative error as
the euclidean distance between the estimation and the ground
truth, normalized by the ground truth. For the speed, the error is computed by averaging on the three components of the relative speed. Finally, for the relative orientation, the error is computed by averaging on the roll pitch and yaw, that define the relative rotation between the two local frames.

In the next subsections, we will present the results obtained
with the closed-form solution provided in section \ref{SectionCFS} on the simulated data
mentioned, with different sensor bias settings. Our goal is to
identify its performance limitations and introduce modifications
to overcome them.

\subsection{Performance without bias}\label{SubSectionPerformanceNoBIAS}

\begin{figure}[htbp]
\begin{center}
\includegraphics[width=.9\columnwidth]{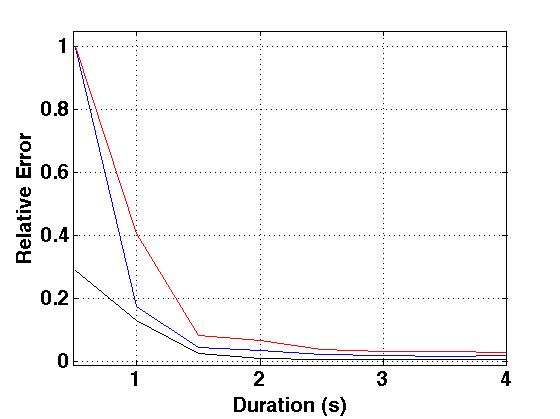}
\caption{Closed-form solution estimations in absence of bias on the inertial measurements. In blue the relative error on the absolute scale, in red on the relative speed and in black on the relative orientation. The two agents observes one each other over
a variable duration of integration.} \label{FigPerfoNoBIAS}
\end{center}
\end{figure}

In Fig. \ref{FigPerfoNoBIAS}, we display the performance of the Closed-Form solution in estimating absolute scale, relative speed and relative orientation. All the values are averaged on $1000$ trials.
Note how the evaluations get better as we increase the
integration time. Note that a robust estimation of the relative orientation
requires a shorter duration of integration than the speed and the
absolute scale.

\subsection{Impact of accelerometer bias on the performance}\label{SubSectionPerformanceBIASAcc}

\begin{figure}[htbp]
\begin{center}
\begin{tabular}{cc}
\includegraphics[width=.4\columnwidth]{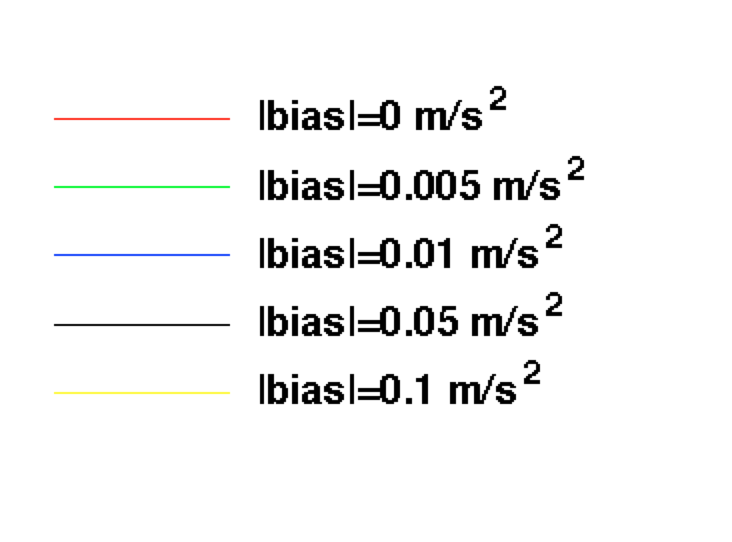}&
\includegraphics[width=.4\columnwidth]{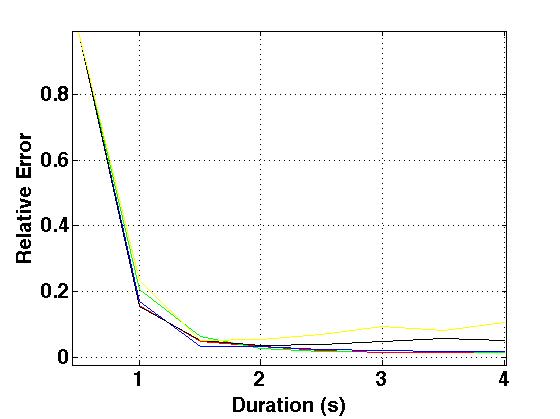}\\
Legend & Absolute Scale\\
\includegraphics[width=.4\columnwidth]{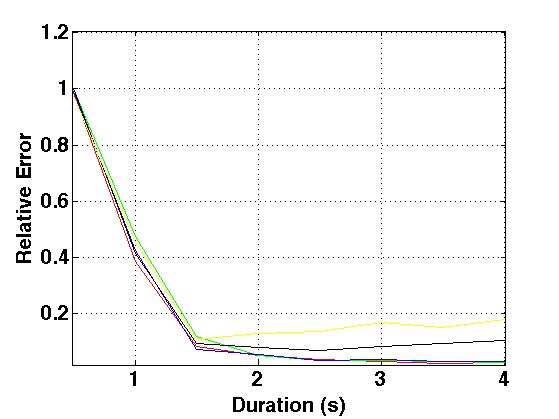}&
\includegraphics[width=.4\columnwidth]{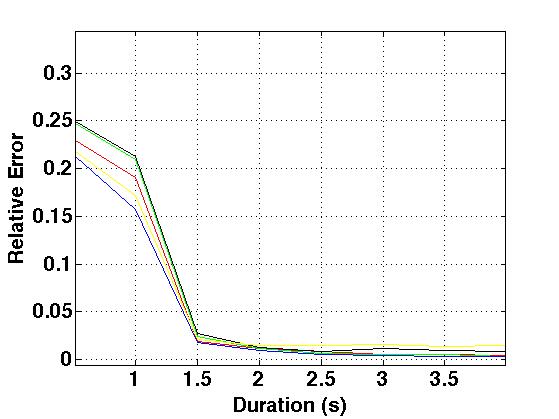}\\
Relative Speed & Relative Orientation
\end{tabular}
\caption{Impact of the accelerometer bias on the performance
of the closed-form solution. The two agents observe one each other over
a variable duration of integration.}
\label{FigPerfoAcc}
\end{center}
\end{figure}

In order to visualize the impact of the accelerometer bias on
the performance, we corrupt the accelerometer measurements
by a bias (Fig. \ref{FigPerfoAcc}).
Despite a high accelerometer bias, the closed-form solution
still provides good results. As in Fig \ref{FigPerfoNoBIAS},  all the values are averaged on $1000$ trials. Note that, even in the case of a bias with magnitude $0.1ms^{-1}$ (yellow line in Fig. \ref{FigPerfoAcc}), the error attains its minimum after $1.5s$ and it is less than $3\%$ for the scale and less than $10\%$ for the relative speed (note that, the larger error on the speed is due to its smaller absolute value, which is $\simeq 0.8 ms^{-1}$, meaning that a $10\%$ error corresponds to an error of $0.08ms^{-1}$).

\subsection{Impact of gyroscope bias on the performance}\label{SubSectionPerformanceBIASGyro}

\begin{figure}[htbp]
\begin{center}
\begin{tabular}{cc}
\includegraphics[width=.4\columnwidth]{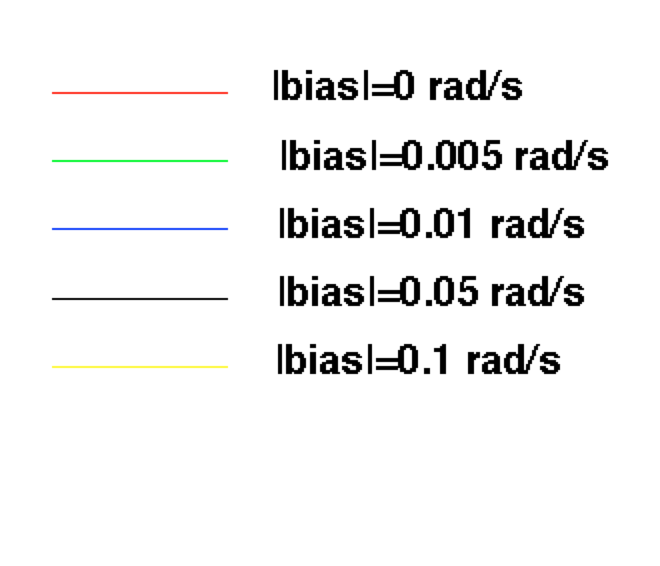}&
\includegraphics[width=.4\columnwidth]{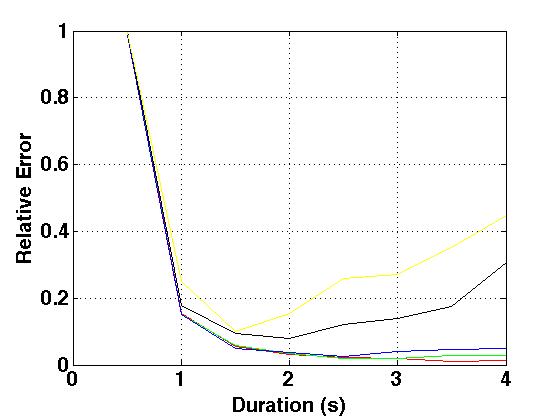}\\
Legend & Absolute Scale\\
\includegraphics[width=.4\columnwidth]{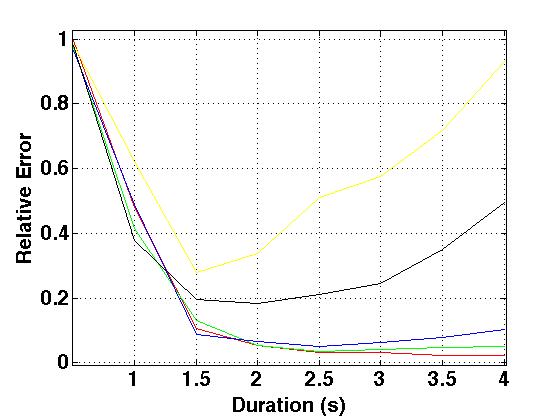}&
\includegraphics[width=.4\columnwidth]{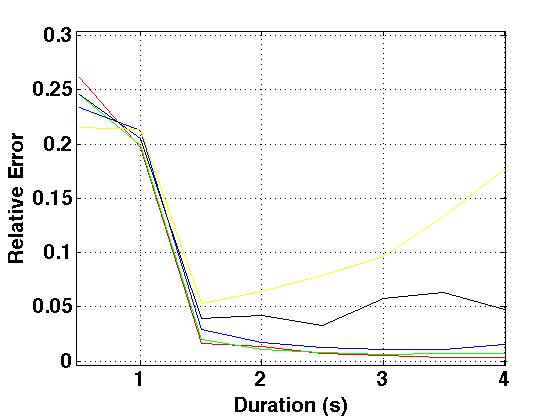}\\
Relative Speed & Relative Orientation
\end{tabular}
\caption{Impact of the gyroscope bias on the performance
of the closed-form solution. The two agents observe one each other over
a variable duration of integration.}
\label{FigPerfoGyro}
\end{center}
\end{figure}

To visualize the impact of the gyroscope bias on the
performance, we corrupt the gyroscope measurements by an
artificial bias (Fig. \ref{FigPerfoGyro}). As in Fig \ref{FigPerfoNoBIAS},  all the values are averaged on $1000$ trials. 
As seen in Fig. \ref{FigPerfoGyro}, the performance becomes very poor in
presence of a bias on the gyroscope and, in practice, the overall
method could only be successfully used with a very precise -
and expensive - gyroscope.

\section{Estimating the Gyroscope Bias ($B^1_\Omega$ and $B^2_\Omega$)}\label{SectionBiasCalibration}

In this section, we propose
an optimization approach to estimate the gyroscope bias using
the closed-form solution.

Since system in (\ref{EquationLinearSystem}) is overconstrained, inverting
it is equivalent to finding the vector $x$ that minimizes the residual $|\Xi x-b|^2$. We define the following cost function:

\begin{equation}\label{EquationCost}
Cost(B)=|\Xi x-b|^2
\end{equation}

\noindent with:

\begin{itemize}

\item $B$ is a vector with six components, which are the components of the bias of the first and the second gyroscope, i.e.,: $B=[B^1_\Omega, ~B^2_\Omega]$.

\item $\Xi$ and $b$ are computed by removing from the measurements provided by the two gyroscopes, the corresponding components of $B$.

\end{itemize}

\noindent By minimizing this cost function, we recover the gyroscope
bias $B$ and the vector $x$. Since our cost function
requires an initialization and is non-convex, the
optimization process can be stuck in local minima. However,
by running extensive simulations we found that the cost
function is convex around the true value of the bias. Hence, we can initialize the optimization process with $B = 0_6$ since
the bias is usually rather small.

\section{Performance Overall Evaluation}\label{SectionExperiments}

\begin{figure}[htbp]
\begin{center}
\begin{tabular}{cc}
\includegraphics[width=.4\columnwidth]{LegendaGyro.png}&
\includegraphics[width=.4\columnwidth]{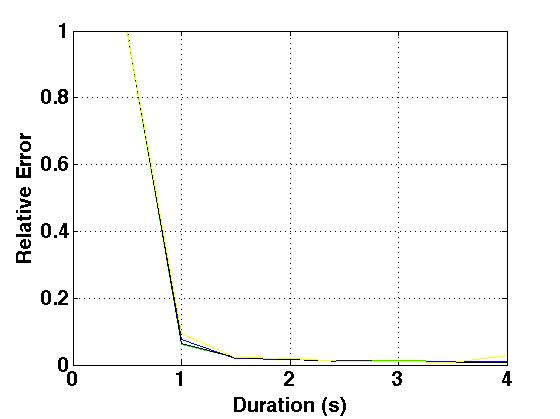}\\
Legend & Absolute Scale\\
\includegraphics[width=.4\columnwidth]{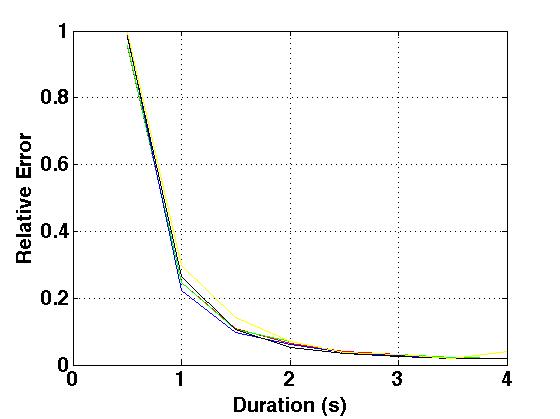}&
\includegraphics[width=.4\columnwidth]{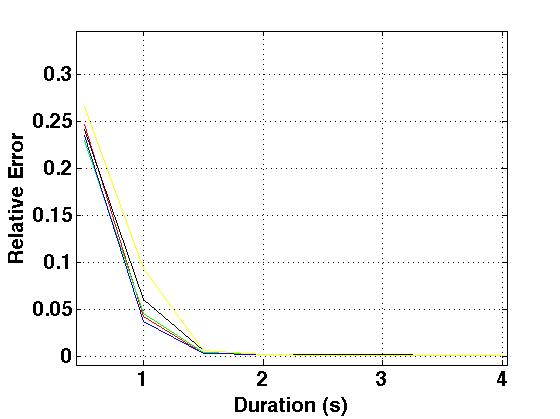}\\
Relative Speed & Relative Orientation
\end{tabular}
\caption{Impact of the gyroscope bias on the performance
of the closed-form solution completed with the bias estimator. The accelerometers are unbiased. The two agents observe one each other over
a variable duration of integration.}
\label{FigPerfoOpt0}
\end{center}
\end{figure}

\begin{figure}[htbp]
\begin{center}
\begin{tabular}{cc}
\includegraphics[width=.4\columnwidth]{LegendaGyro.png}&
\includegraphics[width=.4\columnwidth]{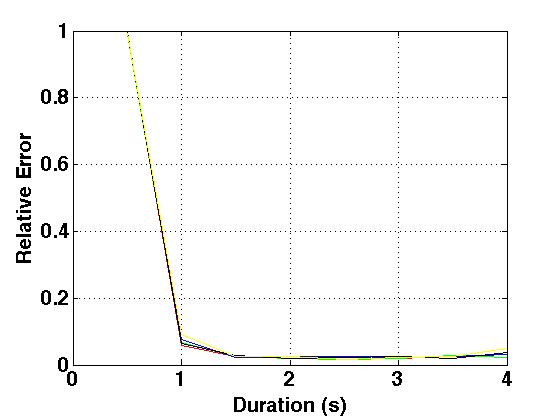}\\
Legend & Absolute Scale\\
\includegraphics[width=.4\columnwidth]{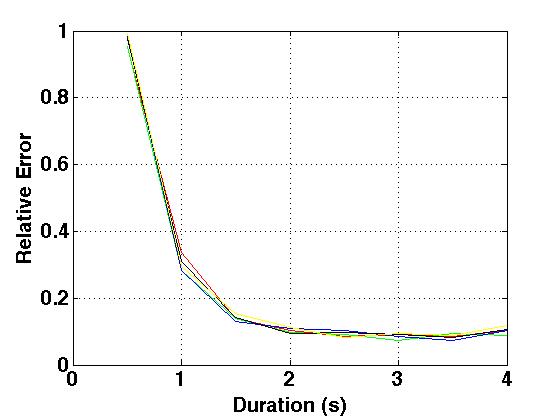}&
\includegraphics[width=.4\columnwidth]{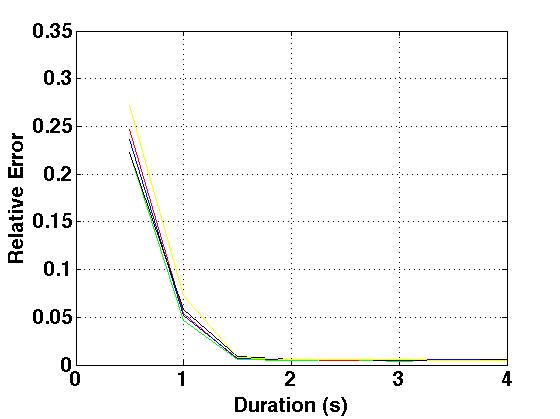}\\
Relative Speed & Relative Orientation
\end{tabular}
\caption{As in Fig. \ref{FigPerfoOpt0} but the magnitude of the accelerometer bias is set to $0.1ms^{-2}$ for both the agents.}
\label{FigPerfoOpt10}
\end{center}
\end{figure}

This section analyzes the performance of the closed form solution completed with the bias estimator introduced in section \ref{SectionBiasCalibration}. The setup is the one described in section \ref{SubSectionSimulationSetup}. Also in this case, the results are averaged on $1000$ trials. We consider the same five values of the bias of the gyroscopes considered in Fig. \ref{FigPerfoGyro}.
Finally, we set the magnitude of accelerometer bias equal to zero (Fig. \ref{FigPerfoOpt0}) and equal to $0.1ms^{-2}$ (Fig. \ref{FigPerfoOpt10}). Fig. \ref{FigPerfoOpt0} shows a performance comparable to the one exhibited in Fig. \ref{FigPerfoNoBIAS}. On the other hand,  Fig. \ref{FigPerfoOpt10} shows a performance better than the one exhibited in Fig. \ref{FigPerfoAcc}. This demonstrates that the effect of the bias has fully overcome. 

%
%
%

\section{Conclusion}\label{SectionConclusion}
In this paper we extended the results presented in \cite{Kai16,TRO12} to the cooperative case. Specifically, the case of two agents was investigated. Each agent was equipped with inertial sensors (accelerometer and gyroscope) and with a monocular camera. By using the monocular camera, each agent can observe the other agent. No additional camera observations (e.g., of external point features in the environment) were considered. All the inertial sensors were assumed to be affected by a bias.
First, the entire observable state was analytically derived. To this regard, we proved that the entire observable state consists of the following independent physical quantities:

\begin{itemize}

\item The position of one of the agents in the local frame of the other agent (this means that the absolute scale is observable).

\item The relative speed between the two agents expressed in the local frame of one of them.

\item The three Euler angles that characterize the rotation between the two local frames attached to the two agents.

\item All the bias that affect the inertial measurements (both the accelerometers and the gyroscopes).

\end{itemize}

\noindent Note that this result holds even in the case when only one of the two agents is equipped with a camera and, very surprisingly, even when this camera is a linear camera, i.e., it only provides the azimuth of the other agent in its local frame.

Then, the paper provided a deterministic solution, able to provide the observable state by only using visual and inertial measurements delivered in a short time interval ($2,3$ seconds).
This solution extended the solution in \cite{TRO12,IJCV14} to the cooperative case. Note that it is possible to retrieve the absolute scale even when no point features are available in the environment.

Following the analysis conducted in \cite{Kai16}, the paper focused on investigating all the limitations that characterize this solution when used in a real scenario. Specifically. the impact of the presence of the bias on the performance of this closed-form solution was investigated. As in the case of a single agent, this performance is significantly  sensitive to the presence of a bias on the gyroscope, while, the presence of a bias on the accelerometer is negligible. A simple and effective method to obtain the gyroscope bias was proposed.
Extensive simulations clearly showed that the proposed method is successful. Note that, it is possible to automatically retrieve the absolute scale and simultaneously calibrate the gyroscopes not only without any prior knowledge (as in \cite{Kai16} for a single agent), but also without external point features in the environment.

Future works will be focused on the analysis of all the singularities that characterize the closed form solution. This will extend the analysis in \cite{IJCV14}. Additionally, we are interested in obtaining a closed form solution in the minimal case of a linear camera. Finally, as it has been done in \cite{TRO12}, we could include the bias on the two accelerometers in the closed form solution even if, due to its slight impact on the performance, we expect that its determination will be not precise.

\appendix

\section{Observability Analysis}\label{AppendixOASystem}

The goal of this appendix is to detect all the observable quantities (or observable modes) of the system defined in section \ref{SectionSystem}. These scalar quantities will be included in a single vector that will be named the {\it observable state}.

We adopt the following notations:

\begin{itemize}

\item $r^1=[x^1, ~y^1, ~z^1]$ and  $r^2=[x^2, ~y^2, ~z^2]$ are the positions of $\mathcal{B}_1$ and $\mathcal{B}_2$ in the global frame;

\item $v^1=[v_x^1, ~v_y^1, ~v_z^1]$ and  $v^2=[v_x^2, ~v_y^2, ~v_z^2]$ are the velocities of $\mathcal{B}_1$ and $\mathcal{B}_2$ in the global frame;

\item $q^1=q_t^1+q_x^1 i+q_y^1 j+q_z^1 k$  and $q^2=q_t^2+q_x^2 i+q_y^2 j+q_z^2 k$ are the two unit quaternions that describe the rotations between the global and the two local frames, respectively\footnote{A quaternion $q=qt+q_xi+q_yj+q_zk$ is a unit quaternion if the product with its conjugate is 1, i.e.: $qq^*=q^*q=(qt+q_xi+q_yj+q_zk)(qt-q_xi-q_yj-q_zk)=(q_t)^2+(q_x)^2+(q_y)^2+(q_z)^2=1$}.

\end{itemize}

\noindent In the sequel, for each vector defined in the $3D$ space, the subscript $q$ will be adopted to denote the corresponding imaginary quaternion. 
For instance, regarding the position of $\mathcal{B}_1$, we have: $r_q^1 =0+x^1~i + y^1~j + z^1 ~k$.

The dynamics of the first/second rigid body are:

\begin{equation}\label{EquationDynamics}
\left[\begin{aligned}
\dot{r}^{1/2}_q &= v^{1/2}_q\\
\dot{v}^{1/2}_q &=q^{1/2}A^{1/2}_q(q^{1/2})^*-g k\\
\dot{q}^{1/2} &= \frac{1}{2}q^{1/2}\Omega^{1/2}_q\\
\end{aligned}
\right.
\end{equation}

\noindent where $g$ is the magnitude of the gravity and $k$ is the fourth fundamental quaternion unit ($k=0+0 ~i + 0~j+1~k$). 

The monocular camera on $\mathcal{B}_1$ provides the position of $\mathcal{B}_2$ in the local frame of $\mathcal{B}_1$, up to a scale. The position of $\mathcal{B}_2$ in the local frame of $\mathcal{B}_1$ is given by the three components of the following imaginary quaternion:

\begin{equation}\label{EquationP1}
p^1_q=(q^1)^*(r_q^2-r_q^1)q^1
\end{equation}

\noindent Hence, the first camera provides the ratios of its components:

\begin{equation}\label{EquationFirstCamera}
h^1\triangleq  [h_u^1,~h_v^1]^T=\left[
\frac{[p^1_q]_x}{[p^1_q]_z},
~\frac{[p^1_q]_y}{[p^1_q]_z} \right]^T
\end{equation}

\noindent where the subscripts $x$, $y$ and $z$ indicate
respectively the $i$, $j$ and $k$ component of the corresponding
quaternion. Similarly, the second camera provides:

\begin{equation}\label{EquationSecondCamera}
h^2\triangleq  [h_u^2,~h_v^2]^T=\left[
\frac{[p^2_q]_x}{[p^2_q]_z},
~\frac{[p^2_q]_y}{[p^2_q]_z} \right]^T
\end{equation}

\noindent where $p^2_q$ is the imaginary quaternion whose three components are the position of $\mathcal{B}_1$ in the local frame of $\mathcal{B}_2$, namely:

\begin{equation}\label{EquationP2}
p^2_q=(q^2)^*(r_q^1-r_q^2)q^2
\end{equation}

We want to obtain the entire observable state. First of all, we characterize our system by the following state:

\begin{equation}\label{EquationState}
X=[r^1~v^1~q^1~r^2~v^2~q^2]^T
\end{equation}

\noindent The dimension of this state is equal to $20$. Actually, the components of this state are not independent. Both $q^1$ and $q^2$ are unit quaternions. In other words, we have: $(q_t^1)^2+(q_x^1)^2+(q_y^1)^2+(q_z^1)^2=(q_t^2)^2+(q_x^2)^2+(q_y^2)^2+(q_z^2)^2=1$.

The dynamics of the state defined in (\ref{EquationState}) are given by (\ref{EquationDynamics}). The observation functions are the four scalar functions $h_u^1 ~h_v^1 ~h_u^2 ~h_v^2$ given by equations (\ref{EquationP1}-\ref{EquationP2}). Additionally, we need to add the two observation functions, which express the constraint that the two quaternions, $q^1$ and $q^2$, are unit quaternions. The two additional observations are:

\begin{equation}\label{EquationQUnit}
h_{const}^{1/2}(X)\triangleq  (q_t^{1/2})^2+(q_x^{1/2})^2+(q_y^{1/2})^2+(q_z^{1/2})^2
\end{equation}

\noindent We investigate the observability properties of this system. Since both the dynamics and the six observations are nonlinear with respect to the state, we use the observability rank condition in \cite{Her77}. The dynamics are affine in the inputs, i.e., they have the expression

\begin{equation}\label{EquationDynamicsAffine}
\dot{X}=f_0(X) + \sum_{i=1}^{12}f_i(X)u_i
\end{equation}

\noindent where $u_i$ are the system inputs, which are the quantities measured by the two IMUs. Specifically, we set:

\begin{itemize}

\item $u_1,u_2,u_3$ the three components of $A^1$;

\item $u_4,u_5,u_6$ the three components of $\Omega^1$;

\item $u_7,u_8,u_9$ the three components of $A^2$;

\item $u_{10},u_{11},u_{12}$ the three components of $\Omega^2$;

\end{itemize}

\noindent Then, by comparing (\ref{EquationDynamics}) with (\ref{EquationDynamicsAffine}) it is immediate to obtain the analytic expression of all the vector fields $f_0,f_1,\cdots f_{12}$; for instance, we have:
{\small
\[
f_0=[ v^1_x, v^1_y, v^1_z, 0, 0, -g, 0, 0, 0, 0, v^2_x, v^2_y, v^2_z, 0, 0, -g, 0, 0, 0, 0]^T
\]
\[
f_4=\frac{1}{2} [0, 0, 0, 0, 0, 0, -q^1_x, q^1_t, q^1_z, -q^1_y, 0, 0, 0, 0, 0, 0, 0, 0, 0, 0]^T
\]
}
\noindent For systems with the dynamics given in (\ref{EquationDynamicsAffine}) the application of the observability rank condition can be automatically done by a recursive algorithm. In particular, this algorithm automatically returns the observable codistribution by computing the Lie derivatives of all the system outputs along all the vector fields that characterize the dynamics (see chapter 1 of \cite{Isi95}). For the specific case, we obtain that the algorithm converges at the third step, i.e., the observable codistribution is the span of the differentials of the previous Lie derivatives up to the second order. In particular, its dimension is $11$ and, a choice of eleven Lie derivatives is: $\mathcal{L}^0h^1_u, ~\mathcal{L}^0h^1_v, ~\mathcal{L}^0h^2_u, ~\mathcal{L}^0h^2_v, ~\mathcal{L}^0h^1_{const}, ~\mathcal{L}^0h^2_{const}, ~\mathcal{L}^1_{f_0}h^1_u, $ $\mathcal{L}^1_{f_0}h^1_v, ~\mathcal{L}^1_{f_0}h^2_u, ~\mathcal{L}^2_{f_0f_0}h^1_u, ~\mathcal{L}^2_{f_0f_1}h^1_u$. Note that the choice of these eleven independent Lie derivatives is not unique. In particular, it is possible to avoid the Lie derivatives of the functions $h^2_u$ and $h^2_v$. Specifically, a possible choice is:
$\mathcal{L}^0h^1_u, ~\mathcal{L}^0h^1_v, ~\mathcal{L}^0h^1_{const}, ~\mathcal{L}^0h^2_{const}, ~\mathcal{L}^1_{f_0}h^1_u, $ $\mathcal{L}^1_{f_0}h^1_v, ~\mathcal{L}^2_{f_0f_0}h^1_u, ~\mathcal{L}^2_{f_0f_1}h^1_u, ~\mathcal{L}^2_{f_0f_7}h^1_u, ~\mathcal{L}^2_{f_0f_8}h^1_u, ~\mathcal{L}^2_{f_0f_7}h^1_v$.
This means that we obtain the same observability properties when only $\mathcal{B}_1$ (or only $\mathcal{B}_2$) is equipped with a camera. In other words, the presence of two cameras does not improve the observability properties with respect to the case of a single camera mounted on one of the two rigid bodies.

Once we have obtained the observable codistribution, the next step is to obtain the observable state. This state has eleven components. Obviously, a possible choice would be the state that contains the previous eleven Lie derivatives. On the other hand, their expression is too complex and it is much more preferable to find an easier state, whose components have a clear physical meaning. By analytically computing the continuous symmetries of our system (i.e., the killing vectors of the previous observable codistribution, \cite{TRO11}), we detect the following independent observable modes:

\begin{itemize}

\item The position of $\mathcal{B}_2$ in the local frame of $\mathcal{B}_1$ (three observable modes);

\item The velocity of $\mathcal{B}_2$ in the local frame of $\mathcal{B}_1$ (three observable modes);

\item The three Euler angles that characterize the rotation between the two local frames  (three observable modes);

\item Trivially, the norm of the two quaternions (two observable modes).

\end{itemize}

\noindent Therefore, we can fully characterize our system by a state whose components are the previous observable modes.

\section{Fundamental Equations}\label{AppendixBasicEquations}

In accordance with the observability analysis carried out in appendix \ref{AppendixOASystem} , we characterize our system by the following state:

\begin{equation}\label{EquationObsState}
S=[R~V~q]^T
\end{equation}

\noindent where $q$ is the unit quaternion that describes the relative rotation between $\mathcal{B}_1$ and $\mathcal{B}_2$.

The imaginary quaternions associated to $R$ and $V$ are:

\begin{equation}\label{EquationX}
R_q=(q^1)^*(r_q^2-r_q^1)q^1
\end{equation}

\begin{equation}\label{EquationV}
V_q=(q^1)^*(v_q^2-v_q^1)q^1
\end{equation}

\noindent and

\begin{equation}\label{EquationQ}
q=(q^1)^*q^2
\end{equation}

\noindent The fundamental equations of the cooperative visual-inertial sensor fusion problem are obtained by differentiating the previous three quantities with respect to time and by using (\ref{EquationDynamics}) in order to express the dynamics in terms of the components of the state in (\ref{EquationObsState}) and the components of $A^1,A^2,\Omega^1,\Omega^2$. After some analytic computation, we obtain:

\begin{equation}\label{EquationObsDynamics}
\left[\begin{aligned}
\dot{R}_q &= \frac{1}{2}(\Omega^1_q)^*R_q + \frac{1}{2} R_q\Omega^1_q + V_q \\
\dot{V}_q &= \frac{1}{2}(\Omega^1_q)^*V_q + \frac{1}{2} V_q\Omega^1_q +q A^2_q q^* - A^1_q\\
\dot{q} &= \frac{1}{2}(\Omega^1_q)^*q + \frac{1}{2} q\Omega^2_q\\
\end{aligned}
\right.
\end{equation}

\noindent As desired, the dynamics of the state is expressed only in terms of the components of the state and the system inputs (the angular speeds and the accelerations of both $\mathcal{B}_1$ and $\mathcal{B}_2$). Finally, the camera observations can be immediately expressed in terms of the state in (\ref{EquationObsState}). The first camera provides the vector $R$ up to a scale. Regarding the second camera, we first need the position of $\mathcal{B}_1$ in the local frame of $\mathcal{B}_2$. The components of this position are the components of the following imaginary quaternion: $-q^*R_qq$. The second camera provides this position up to a scale.

\vskip .2cm

In the last part of this appendix we provide the same equations, without using quaternions. We characterize our system by the two 3D vectors $R$ and $V$, as before. Instead of the quaternion $q$, we use the matrix $O$ that characterizes the rotation between the two local frames. From (\ref{EquationObsDynamics}) it is immediate to obtain the dynamics of this state. They are the equations in (\ref{EquationObsDynamicsNoq}).

\section{Observability with bias}\label{AppendixObservabilityBIAS}

We analytically obtain the observability properties of the system defined by the state in (\ref{EquationState}), the dynamics in (\ref{EquationDynamicsBIAS}) and the three observations in (\ref{EquationFirstCamera}) and (\ref{EquationQUnit}). Since both the dynamics and the observations are nonlinear with respect to the state, we use the observability rank condition in \cite{Her77}. The dynamics are affine in the inputs, i.e., they have the expression

\begin{equation}\label{EquationDynamicsAffine}
\dot{X}=f_0(X) + \sum_{i=1}^{12}f_i(X)u_i
\end{equation}

\noindent where $u_i$ are the system inputs, which are the quantities measured by the two IMUs. Specifically, we set:

\begin{itemize}

\item $u_1,u_2,u_3$ the three components of $\Omega^1$;

\item $u_4,u_5,u_6$ the three components of $A^1$;

\item $u_7,u_8,u_9$ the three components of $\Omega^2$;

\item $u_{10},u_{11},u_{12}$ the three components of $A^2$;

\end{itemize}

\noindent Then, by comparing (\ref{EquationDynamicsBIAS}) with (\ref{EquationDynamicsAffine}) it is immediate to obtain the analytic expression of all the vector fields $f_0,f_1,\cdots f_{12}$; for instance, we have:
{\small
\[
f_4 =[0,0,0,q_t^2 + q_x^2 - q_y^2 - q_z^2, 2q_tq_z + 2q_xq_y,2q_xq_z - 2q_tq_y,0,0,0,0,0_{12}]^T
\]
\[
f_7 =[0,-R_z,R_y,0,-V_z,V_y,q_x/2,-q_t/2,q_z/2,-q_y/2,0_{12}]^T
\]
}
\noindent For systems with the dynamics given in (\ref{EquationDynamicsAffine}) the application of the observability rank condition can be automatically done by a recursive algorithm. In particular, this algorithm automatically returns the observable codistribution by computing the Lie derivatives of all the system outputs along all the vector fields that characterize the dynamics (see chapter 1 of \cite{Isi95}). For the specific case, we obtain that the algorithm converges at the fourth step, i.e., the observable codistribution is the span of the differentials of the previous Lie derivatives up to third order. In particular, its dimension is $22$ meaning that all the state components are observable. A choice of $22$ Lie derivatives is: 
{\tiny
$\mathcal{L}^0h_u, ~\mathcal{L}^0h_{const}, ~\mathcal{L}^1_{f_0}h_u,~\mathcal{L}^1_{f_7}h_u, 
~\mathcal{L}^2_{f_0f_0}h_u,~\mathcal{L}^2_{f_0f_1}h_u,~\mathcal{L}^2_{f_0f_4}h_u,~\mathcal{L}^2_{f_0f_5}h_u,$ $\mathcal{L}^2_{f_0f_7}h_u,~\mathcal{L}^2_{f_0f_8},~\mathcal{L}^2_{f_0f_9}h_u,~\mathcal{L}^3_{f_0f_0f_0}h_u,~\mathcal{L}^3_{f_0f_0f_1}h_u,~\mathcal{L}^3_{f_0f_0f_2}h_u,~\mathcal{L}^3_{f_0f_0f_4}h_u,$
$~\mathcal{L}^3_{f_0f_0f_7}h_u,~\mathcal{L}^3_{f_0f_0f_8}h_u,~\mathcal{L}^3_{f_0f_0f_{10}}h_u,~\mathcal{L}^3_{f_0f_4f_0}h_u,
$
$~\mathcal{L}^3_{f_0f_5f_0}h_u,~\mathcal{L}^3_{f_0f_0f_0}h_v,~\mathcal{L}^3_{f_0f_0f_8}h_v$
}. Note that the choice of these $22$ independent Lie derivatives is not unique. In particular, it is possible to avoid the Lie derivatives of the functions $h_v$. Specifically, in the previous choice, only the last two Lie derivatives are Lie derivatives of the function $h_v$. It is possible to avoid these two functions. On the other hand, in this case we need to include fourth order Lie derivatives of $h_u$. For instance, we can replace the last two functions with {\tiny $~\mathcal{L}^4_{f_0f_0f_0f_0}h_u,~\mathcal{L}^4_{f_0f_0f_0f_8}h_u$}.
This means that we obtain the same observability properties when the first agent is equipped with a linear camera able to only provide the azimuth of the second agent in its local frame.

\end{document}